\documentclass[sigconf]{acmart}
\pdfoutput=1
\pdfoutput=1
\pdfoutput=1
\pdfoutput=1
\pdfoutput=1
\AtBeginDocument{%
  \providecommand\BibTeX{{%
    \normalfont B\kern-0.5em{\scshape i\kern-0.25em b}\kern-0.8em\TeX}}}

\pdfoutput=1
\copyrightyear{2022}
\acmYear{2022}
\setcopyright{acmcopyright}\acmConference[ICMR '22]{Proceedings of the 2022 International Conference on Multimedia Retrieval}{June 27--30, 2022}{Newark, NJ, USA}
\acmBooktitle{Proceedings of the 2022 International Conference on Multimedia Retrieval (ICMR '22), June 27--30, 2022, Newark, NJ, USA}
\acmPrice{15.00}
\acmDOI{10.1145/3512527.3531375}
\acmISBN{978-1-4503-9238-9/22/06}

\usepackage{multirow}
\usepackage{multicol}
\usepackage{hhline}
\usepackage{color, colortbl}
\definecolor{yellow}{rgb}{1, 0.95, 0.51}
\definecolor{pink}{rgb}{0.98, 0.8, 0.91}
\definecolor{applegreen}{rgb}{0.55, 0.71, 0.0}

\usepackage{CJKutf8}
\usepackage{pinyin} 

\acmSubmissionID{110}

\begin{document}

\title{Cross-lingual Adaptation for Recipe Retrieval with Mixup}
\author{Bin Zhu}
\affiliation{%
  \institution{University of Bristol}
}
\email{bin.zhu@bristol.ac.uk}

\author{Chong-Wah Ngo}
\affiliation{%
  \institution{Singapore Management University}
}
\email{cwngo@smu.edu.sg}

\author{Jingjing Chen}
\affiliation{%
  \institution{Fudan University}
}
\email{chenjingjing@fudan.edu.cn}

\author{Wing-Kwong Chan}
\affiliation{%
  \institution{City University of Hong Kong}
}
\email{wkchan@cityu.edu.hk}

\begin{abstract}
       Cross-modal recipe retrieval has attracted research attention in recent years, thanks to the availability of large-scale paired data for training. Nevertheless, obtaining adequate recipe-image pairs covering the majority of cuisines for supervised learning is difficult if not impossible.
   By transferring knowledge learnt from a data-rich cuisine to a data-scarce cuisine, domain adaptation sheds light on this practical problem.
   Nevertheless, existing works assume recipes in source and target domains are mostly originated from the same cuisine and written in the same language.
   This paper studies unsupervised domain adaptation for image-to-recipe retrieval, where recipes in source and target domains are in different languages. Moreover, only recipes are available for training in the target domain. A novel
   recipe mixup method is proposed to learn transferable embedding features between the two domains. %
   Specifically, recipe mixup produces mixed recipes to form an intermediate domain by discretely exchanging the section(s) between source and target recipes. To bridge the domain gap, recipe mixup loss is proposed to enforce the intermediate domain to locate in the shortest geodesic path between source and target domains in the recipe embedding space.
    By using Recipe 1M dataset as source domain (English) and Vireo-FoodTransfer dataset as target domain (Chinese), empirical experiments verify the effectiveness of recipe mixup for cross-lingual adaptation in the context of image-to-recipe retrieval. 
\end{abstract}

\ccsdesc[500]{Information systems~Multimedia and multimodal retrieval}

\keywords{recipe retrieval, mixup, cross-lingual, domain adaptation}

\maketitle

\section{Introduction}
\label{sec:intro}
The prevalence of cooking sharing websites accumulates plenty of recipe-image pairs for training cross-modal deep networks. Cross-modal recipe retrieval, i.e., searching the recipe by using the food image as query, becomes feasible with these networks~\cite{JNE,AdaMine,jingjing18,R2GAN,ACME,MCEN,salvador2021revamping,sugiyama2021cross, zhu2021learning}.
With an image-to-recipe retrieval system, applications such as food recognition~\cite{bossard2014food,jingjing16,jingjing17, chen2020study}, food recommendation~\cite{min2019food, gao2019hierarchical}, food image generation~\cite{pan2020chefgan, CookGAN}, nutrition estimation and food logging~\cite{kitamura2009foodlog, beijbom2015menu, ming2018food} will be benefited.

To generalize recipe retrieval models for different cuisines, cross-modal domain adaptation is explored in~\cite{CCFT}. Leveraging on the abundant recipe-image pairs in a source domain, the aim is to learn an aligned embedding space for retrieving the recipes in a target domain with few or even no training pairs of data.
As cross-domain model adaptation is capable of learning generalizable representation, the burden of data collection, especially for data-scare cuisine, can be alleviated.
Existing works only study cross-modal retrieval in the context of monolingual recipes, e.g., English~\cite{JNE,jingjing18,R2GAN,ACME, MCEN}, Chinese~\cite{jingjing16,CCFT} and Japanese~\cite{kikuta2017approaches}.
Nevertheless, recipes are naturally cross-lingual in the open world since the recipes of a cuisine are usually written in a mother tongue language. Transferring knowledge across languages is also becoming an issue for cross-modal recipe retrieval. 

\begin{figure*}
  \centering
  \includegraphics[width=0.86\textwidth]{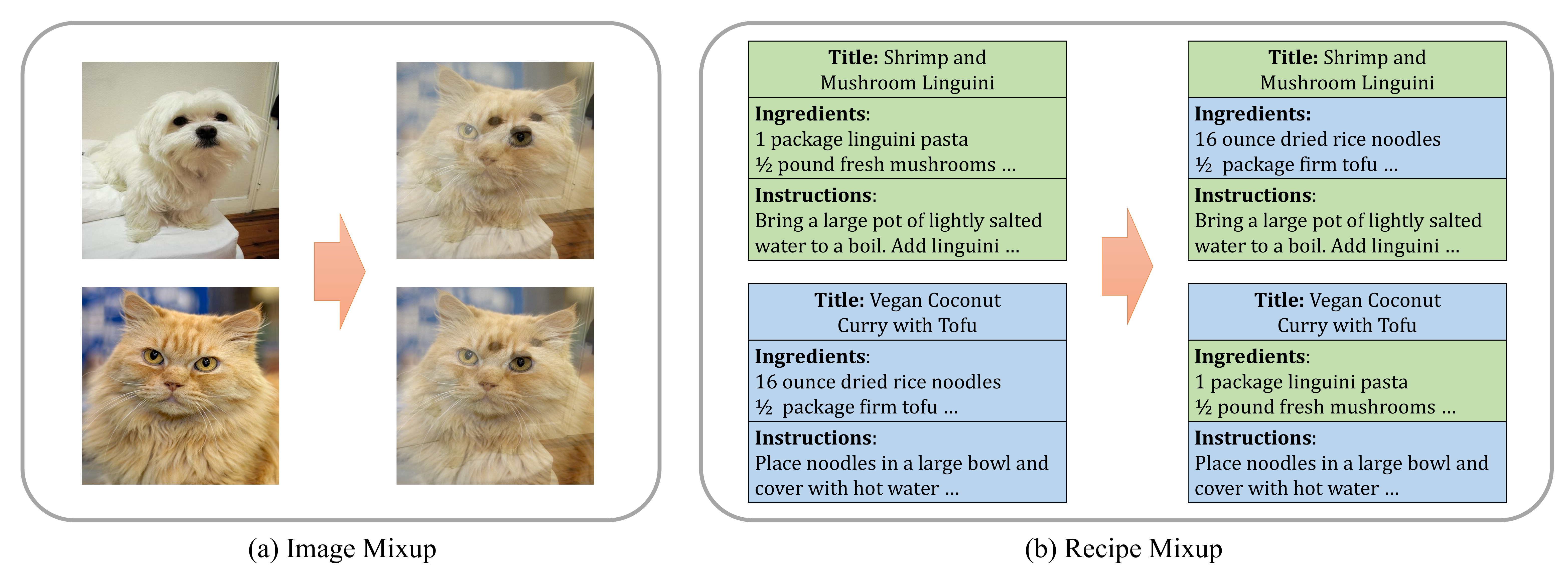}
  \caption{Examples of mixup. (a) Image mixup~\cite{mixup18, domainmixup} constructs new examples by linearly interpolating two random image samples for data augmentation. (b) The proposed recipe mixup produces mixed recipes by discretely exchanging recipe section(s) (e.g., ingredient section) between source and target recipes for domain augmentation.}
  \label{fig:motivation}
\end{figure*}

This paper studies the problem of unsupervised domain adaptation in the context of image-to-recipe retrieval. 
Specifically, given a source domain with abundant recipe-image pairs and a target domain with only recipes, the objective is to transfer knowledge from source to target domains for cross-modal recipe retrieval, where the recipes between domains are written in different languages. Due to missing of paired data in the target domain for performing supervised learning, this task is considered to be unsupervised domain adaptation~\cite{pan2009survey}.
In general, the domain gap stems from the differences in languages as well as the ingredient usage and cooking styles among cuisines.
The nature of recipes makes this task more challenging than other cross-lingual problems, for instance, cross-lingual sentiment classification~\cite{zhou2016cross,chen2018adversarial} and cross-language image-sentence matching~\cite{kim2020mule}. For example, in~\cite{kim2020mule}, the multi-lingual semantic alignment between captions written in different languages is assumed. However, the multi-lingual recipes in this paper are not linguistically or semantically aligned. 

To address these problems, we propose a novel recipe mixup method. Inspired by image mixup~\cite{mixup18,domainmixup}, which constructs new examples by linear interpolation with a ratio between two random images for data augmentation (Figure~\ref{fig:motivation} (a)), recipe mixup is proposed to produce mixed recipes by discretely exchanging section(s) between source and target recipes for domain augmentation (Figure~\ref{fig:motivation} (b)). As a recipe usually consists of three sections including title, ingredients and instructions, recipe mixup can be performed effortlessly by different section combinations of two recipes. For example, as shown in Figure~\ref{fig:motivation} (b), given two recipes from source and target domains, if the ingredient section is exchanged, two symmetric mixed recipes can be obtained. Specifically, the title and instructions are kept while ingredients are swapped between recipes. Intuitively, the mixed recipes share partial content with source and target domains, forming an intermediate domain~\cite{intermediate2013} to bridge the domain gap.
It has been shown that such an intermediate domain should locate in the shortest geodesic path between source and target domain on Grassmann manifold for positive knowledge transfer~\cite{intermediate2013, intermediate2019, intermediate2021}. Based on these prior studies, recipe mixup loss is proposed 
to directly minimize the extra domain shift introduced by the intermediate domain compared with the shortest geodesic path. 
Together with cross-modal learning, ingredient recognition from image embedding, image generation from recipe embedding, as well as adversarial learning for feature alignment between domains, the proposed recipe mixup method outperforms baseline models with a large margin using Recipe 1M~\cite{JNE} as source domain (English) and Vireo-FoodTransfer~\cite{CCFT} as target domain (Chinese).

\section{Related Work}
\subsection{Cross-modal Recipe Retrieval}
The goal of cross-modal recipe retrieval is to retrieve the relevant recipes of a query dish image. 
 Most existing works~\cite{JNE,AdaMine,jingjing18,R2GAN,ACME,MCEN,sugiyama2021cross, salvador2021revamping} focus on learning similarity measurement between recipe and image in a common embedding space.
The efforts are ranged from exploration of attention mechanism~\cite{jingjing18, MCEN},
sample mining~\cite{AdaMine,R2GAN,ACME}, generative adversarial nets~\cite{R2GAN,ACME,sugiyama2021cross} to transformers~\cite{salvador2021revamping, vaswani2017attention, zhang2021token}. 
Nevertheless, the foundation of these works is the assumption of large-scale datasets with recipe-image pairs~\cite{JNE} for model training. The long-tail cuisine effect, specifically the model that is overly dominated by the training pairs from popular cuisines, is ignored.
Consequently, these works are incapable to generalize well if a query image is drawn from a data-scare cuisine.

The limitation is partially addressed by cross-domain cross-modal food transfer (CCFT)~\cite{CCFT}, which is the first work to study cross-domain adaptation for recipe retrieval. Nevertheless, the focus of ~\cite{CCFT} is to address the problem of incomplete recipe transfer. This is different from our paper, where the recipes are complete with three sections but the languages are different. With this intuitively larger domain gap due to different modalities and languages, the contribution of this paper is to propose a parameter-free mixup technique for domain adaptation.

\subsection{Domain Adaptation}
Domain adaptation has been extensively studied in visual domain~\cite{patel2015visual}, which aims to transfer knowledge from a source domain to a target domain. To deal with the domain gap, one typical line of works is to explicitly measure domain discrepancy by metrics, such as maximum mean discrepancy (MMD)~\cite{MMD,long2015learning}. 
Another line of works is to introduce a domain classifier to adversarially learn domain-invariant features~\cite{DANN, tzeng2017adversarial, cao2018partial2, zhang2018importance, hoffman2018cycada}. 
In addition, the idea of mixup is explored in~\cite{domainmixup, mao2019virtual} by conducting linearly interpolation of both image-level and feature-level mixup for domain adaptation. In contrast, the mixup in this paper is conducted on the highly structured but free-form written recipes in a discretely non-linear way for cross-lingual cross-modal adaptation.

Multi-modal~\cite{qi2018unified, huang2018mhtn} and cross-lingual~\cite{chen2018adversarial, chen2019multi, kim2020mule} transfers are also explored in the literature.
Nevertheless, most works study either cross-modal adaptation without considering different languages~\cite{qi2018unified, huang2018mhtn}, or cross-lingual adaptation for single text modality~\cite{chen2018adversarial, chen2019multi}.   
This paper shares similar spirit with~\cite{kim2020mule} which considers both cross-lingual and cross-modal adaptation.
The aim of \cite{kim2020mule} is to learn a multi-lingual embedding space for image-sentence matching. A shared language embedding is learnt by a language-specific
fully-connected layer.
Different from~\cite{kim2020mule}, this paper does not assume that the multi-lingual descriptions are aligned for model learning. Our assumption is that there exists an intermediate domain that bridges the domain gap while the instance-level alignment between recipes of different domains is unknown.

\begin{figure*}
  \centering
  \includegraphics[width=0.96\textwidth]{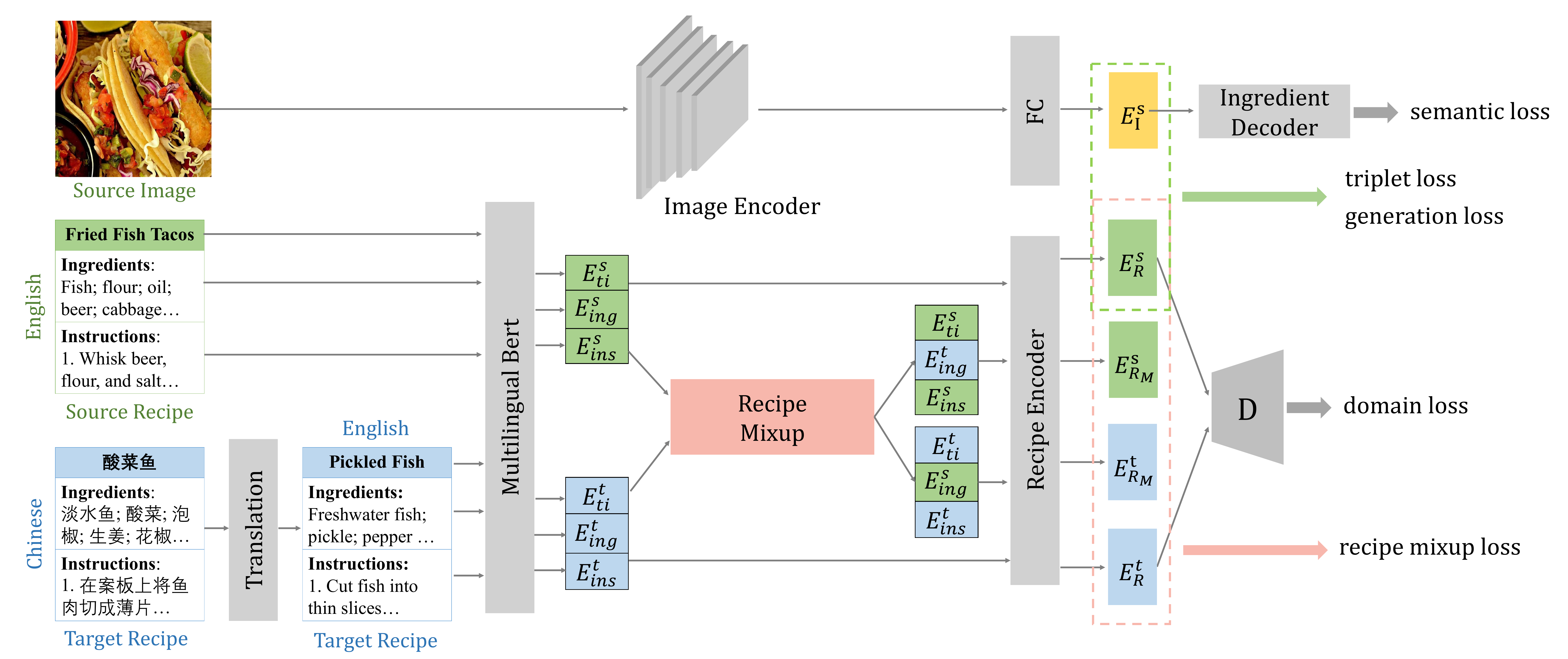}
  \caption{Overview of the proposed model architecture. The recipe mixup block aims to produce mixed recipes to construct intermediate domain by exchanging section(s) between source and target recipes. Ingredient section exchange is used as an example for illustration.}
  \label{fig:model}
\end{figure*}

\begin{figure}
  \centering
  \includegraphics[width=0.48\textwidth]{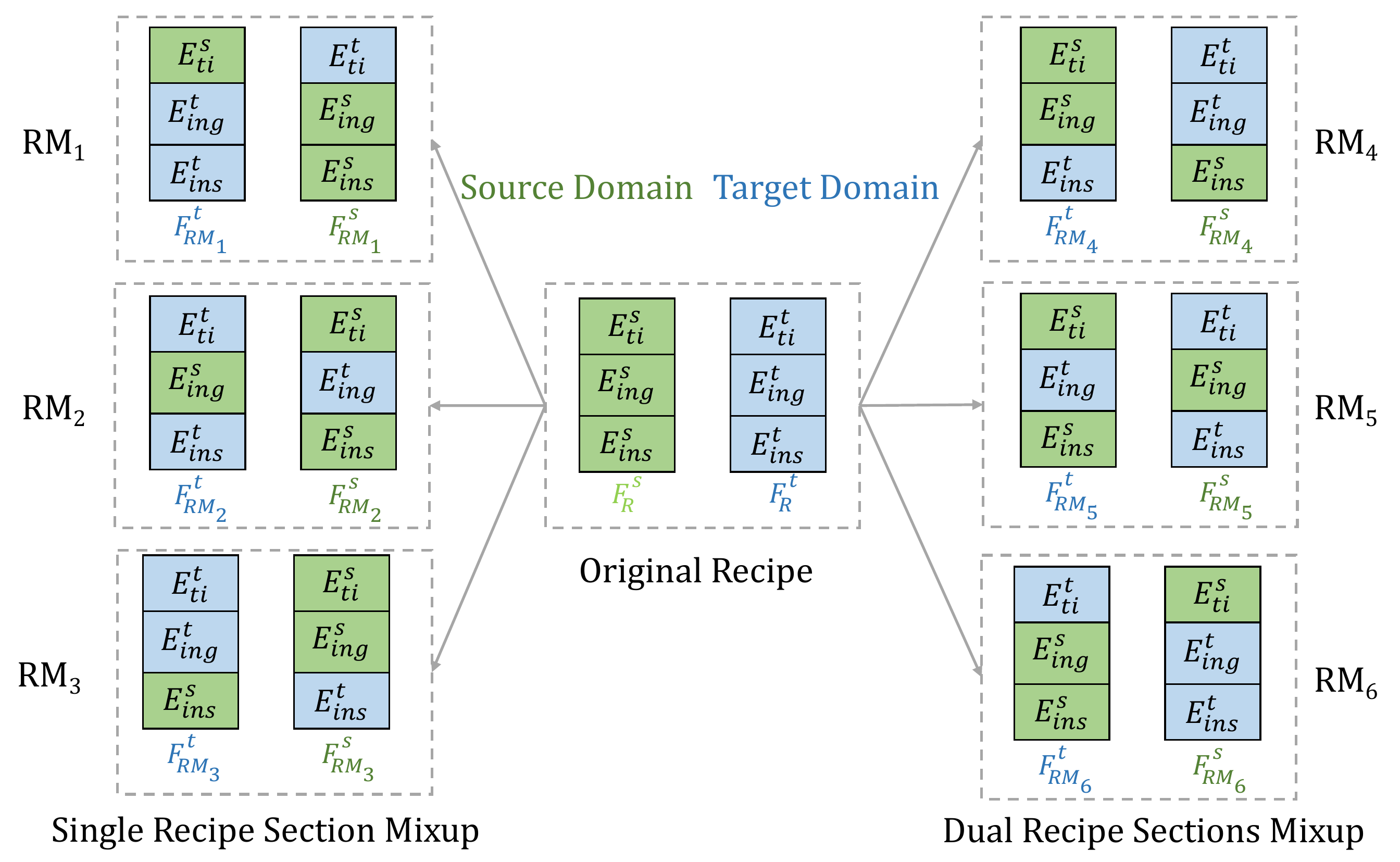}
  \caption{Six kinds of recipe mixup (RM) strategies. RM$_1$, RM$_2$ and RM$_3$ in the left side correspond to one recipe section exchange while RM$_4$, RM$_5$ and RM$_6$ in the right side perform two recipe sections exchange.}
  \label{fig:ss}
\end{figure}

\section{Method}

\subsection{Problem Definition}
Suppose we have a source domain $\mathcal{D}^s={\{(r_i^s, v_i^s)\}}_{i=1}^{N^s}$ with $N^s$ recipe-image pairs and a target domain $\mathcal{D}^t={\{(r_j^t)\}}_{j=1}^{N^t}$ with only $N^t$ recipes, where $r^s_i$ and $r^t_j$
refer to source and target recipes respectively, $v^s_i$ is a source image corresponding to $r^s_i$. 
Leveraged on paired source recipe-image pairs and target recipes for model learning, the goal is to retrieve the corresponding recipes using unseen target images as queries. Note that we assume the source and target recipes are written in different languages, for example, $r^s$ can be written in English while $r^t$ is in Chinese.

\subsection{Model Architecture}
Figure~\ref{fig:model} depicts the overview architecture of the proposed model. Following other cross-lingual works~\cite{nie2010cross,zhou2016cross}, the target recipes are first translated to the source language by machine translation~\cite{MT}. Both source and translated recipes
are subsequently fed into the pre-trained multilingual Bert model~\footnote{https://github.com/google-research/bert/blob/master/multilingual.md}~\cite{Bert} to extract features for each recipe section respectively, i.e., title,
ingredients and instructions. 
To be specific,
the title, each ingredient with quantity and unit (e.g., a teaspoon of sugar, 50 grams pork), and each instruction written as a sentence, are fed into Bert model respectively and transformed to fixed-length vectors. 
The output source and target recipe Bert features are denoted as $F^s_R=\{E_{ti}^s,E_{ing}^s,E_{ins}^s\}$ and $F^t_R=\{E_{ti}^t,E_{ing}^t,E_{ins}^t\}$ respectively,
where $E_{ti}, E_{ing}$ and $E_{ins}$ represent title, ingredient and instruction features.

The key component of our model is recipe mixup block (Section~\ref{subsec:SS}), which aims to
produce mixed recipe features to form an intermediate domain by discretely exchanging section(s) between source and target recipes. Together with $F^s_R$ and $F^t_R$, the mixed recipe features are passed to the recipe encoder to obtain the final
recipe embeddings, including source recipe embedding $E^s_R$, target recipe embedding $E^t_R$, source mixed recipe embedding $E^s_{R_M}$ and target mixed recipe embedding $E^t_{R_M}$. Meanwhile, the source image embedding $E^s_I$ is extracted from the image encoder and mapped into the common embedding space with recipes. As recipe-image pairs are available in the source domain, similar to other works in cross-modal recipe retrieval~\cite{jingjing18, R2GAN, ACME, CCFT}, triplet loss is adopted for common space learning between recipe and image.
The triplet loss is defined as follows:

\begin{equation}
  \begin{aligned}
    \mathcal{L}_{tri} = [d(E^s_a,E^s_p)-d(E^s_a,E^s_n) + \alpha]_+,
  \end{aligned}
  \label{eq:tri}
\end{equation}
where $d(\cdot,\cdot)$ is a distance function measured by cosine similarity, $E^s_a$, $E^s_p$ and $E^s_n$ are the anchor, positive and negative embeddings in the source domain respectively. The parameter $\alpha$ is the margin. 

To align the recipe embeddings between source and target domains, a domain discriminator $D$
is employed to distinguish whether the input recipe embedding comes from source or target domain~\cite{DANN, CCFT}. 
The adversarial domain loss $\mathcal{L}_{adv}$ is defined as follows:

\begin{equation}
  \begin{aligned}
     \mathcal{L}_{adv} = &\mathbb{E}_{E^s_{R}\sim p_{R}^s}[\log{D(E_{R}^s)}]\\
     &+\mathbb{E}_{E_{R}^t\sim p_{R}^t}[\log{(1-D(E_{R}^t)}].
  \end{aligned}
  \label{eq:adv}
\end{equation}

Similar to~\cite{CCFT}, semantic loss $\mathcal{L}_{sem}$ and image generation loss $\mathcal{L}_{gen}$ are also employed to better capture semantic information. Specifically, the source image embedding $E^s_I$ is fed into
an ingredient decoder to predict the ingredients of the food image. The ingredient decoder is a multi-label classifier and trained using cross-entropy loss, i.e., $\mathcal{L}_{sem}$. The source recipe embedding $E^s_R$ is conditioned to reconstruct the
food image. $\mathcal{L}_{gen}$ is computed by the generation error between the reconstructed and real images.

The overall objective function of the proposed method is defined as follows:

\begin{equation}
  \begin{aligned}
      \mathcal{L} = \mathcal{L}_{tri} + \lambda_1\mathcal{L}_{rm} + \lambda_2\mathcal{L}_{adv} + \lambda_3(\mathcal{L}_{sem} + \mathcal{L}_{gen}),
  \end{aligned}
  \label{eq:loss}
\end{equation}
where $\mathcal{L}_{rm}$ denotes recipe mixup loss (Section~\ref{subsec:SS}). The hyper-parameters $\lambda_1$, $\lambda_2$ and $\lambda_3$ balance the importance of the losses.

\subsection{Recipe Mixup (RM)}
\label{subsec:SS}
The key idea of the proposed recipe mixup block is to produce mixed recipes by exchanging recipe section(s) between source and target domains. As the translation and multilingual Bert models are frozen during training, it is equivalent to exchange sections of recipe Bert features between $F^s_R$ and $F^t_R$ in practice. 
Specifically, as shown in Figure~\ref{fig:ss}, given a batch of source and target recipe features, six kinds of recipe mixup strategies are derived by either exchanging one or two sections
between $F^s_R$ and $F^t_R$. If one section is exchanged, three groups of single recipe section mixup features can be obtained as shown in the left side of Figure~\ref{fig:ss}, i.e., RM$_1$, RM$_2$ and RM$_3$ corresponding to the exchange of title, ingredients
and instructions, respectively. On the other hand, if two sections are exchanged, three groups of dual recipe sections mixup features are derived as shown in the right side of Figure~\ref{fig:ss}, i.e., RM$_4$, RM$_5$ and RM$_6$. As the source and target recipe mixups are symmetric, each kind of recipe mixup produces a pair of mixed recipe features, denoted as $\{(F^s_{R_{M_i}}, F^t_{R_{M_i}})\}_{i=1}^6$, where $i$ refers to the number of recipe mixup.
To maintain the consistence of the mixed recipes and reduce the extra domain shift from intermediate domain~\cite{intermediate2021}, 
identical recipe mixup strategy (i.e., one of the six recipe mixup strategies) is employed to form the intermediate domain at one time.

The original and mixed Bert recipe features are subsequently transformed to recipe embeddings by the recipe encoder, i.e., $E_R^s$, $E_R^t$, $E_{R_M}^s$ and $E_{R_M}^t$. Inspired by~\cite{intermediate2013,intermediate2019, intermediate2021}, intermediate domain should lie in the shortest geodesic path between source and target domains on Grassmann manifold. In other words, the sum of the distances between intermediate domain with source and target domains should be equal to the distance between source and target domains, i.e., $d(P^s, P^{inter}) + d(P^t, P^{inter}) = d(P^s, P^t)$, where $d(\cdot,\cdot)$ is a distance measurement between two distributions. $P^s$, $P^t$ and $P^{inter}$ refer to data distribution of source, target and intermediate domains respectively. Violation of the constraint, i.e., $d(P^s, P^{inter}) + d(P^t, P^{inter}) > d(P^s, P^t)$, would introduce ``extra domain shift'', which is more likely to exert negative transfer. Our solution is to minimize the extra domain shift in the embedding space. Following~\cite{intermediate2021}, L2 norm is employed to measure the distribution distance between two domains. As we have two symmetric source and target mixed recipes in each kind of recipe mixup, intermediate domain can be formed with only source mixed recipes ($F^s_{R_{M_i}}$), only target mixed recipes ($F^t_{R_{M_i}}$) or both source and target mixed recipes ($F^s_{R_{M_i}}, F^t_{R_{M_i}}$). The recipe mixup losses corresponding to these three cases are defined as follows: 

\begin{equation}
  \begin{aligned}
    \mathcal{L}_{rm}^s = \underbrace{\left \|E_R^s-E^s_{R_M}\right \|_2+\left \|E_R^t-E^s_{R_M}\right \|_2 - \left \|E_R^s-E^t_{R}\right \|_2}_{\text{extra domain shift from source mixed recipes}},
  \end{aligned}
  \label{eq:rm__source_l2}
\end{equation}

\begin{equation}
  \begin{aligned}
    \mathcal{L}_{rm}^t = \underbrace{\left \|E_R^s-E^t_{R_M}\right \|_2+\left \|E_R^t-E^t_{R_M}\right \|_2-\left \|E_R^s-E^t_{R}\right \|_2}_{\text{extra domain shift from target mixed recipes}},
  \end{aligned}
  \label{eq:rm__target_l2}
\end{equation}
  
\begin{equation}
  \begin{aligned}
    \mathcal{L}_{rm}^{st} = \frac{1}{2}(\mathcal{L}_{rm}^s + \mathcal{L}_{rm}^t).
  \end{aligned}
  \label{eq:rm_l2}
\end{equation}  
  
Note that if source and target mixed recipes are considered independently, the source single recipe section mixup is equivalent to the target dual recipe sections mixup in practice, and vice versa. For instance, as shown in Figure~\ref{fig:ss}, $F^s_{RM_1}$ in RM$_1$ is the same to the $F^t_{RM_6}$ in RM$_6$. Hence, $\mathcal{L}^s_{rm}$  and $\mathcal{L}^t_{rm}$ are somehow interchangeable. Only the results of $\mathcal{L}^s_{rm}$ are reported to avoid redundancy.

\section{Experiments}
\subsection{Experiment Settings}
\label{subsec:exp_ing}
\textbf{Datasets.} The experiments are conducted on Recipe 1M~\cite{JNE} and Vireo-FoodTransfer~\cite{CCFT} datasets, which involve two most widely used languages: English and Chinese respectively.
Recipe 1M contains 341,421 English recipe-image pairs in total, with 4,102 unique ingredients. Vireo-FoodTransfer consists of 70,985 Chinese recipe-images pairs with 1,635 ingredients.
In the experiments, Recipe 1M and Vireo-FoodTransfer are regarded as the source and target domains respectively. 

\begin{figure*}
  \centering
  \includegraphics[width=0.9\textwidth]{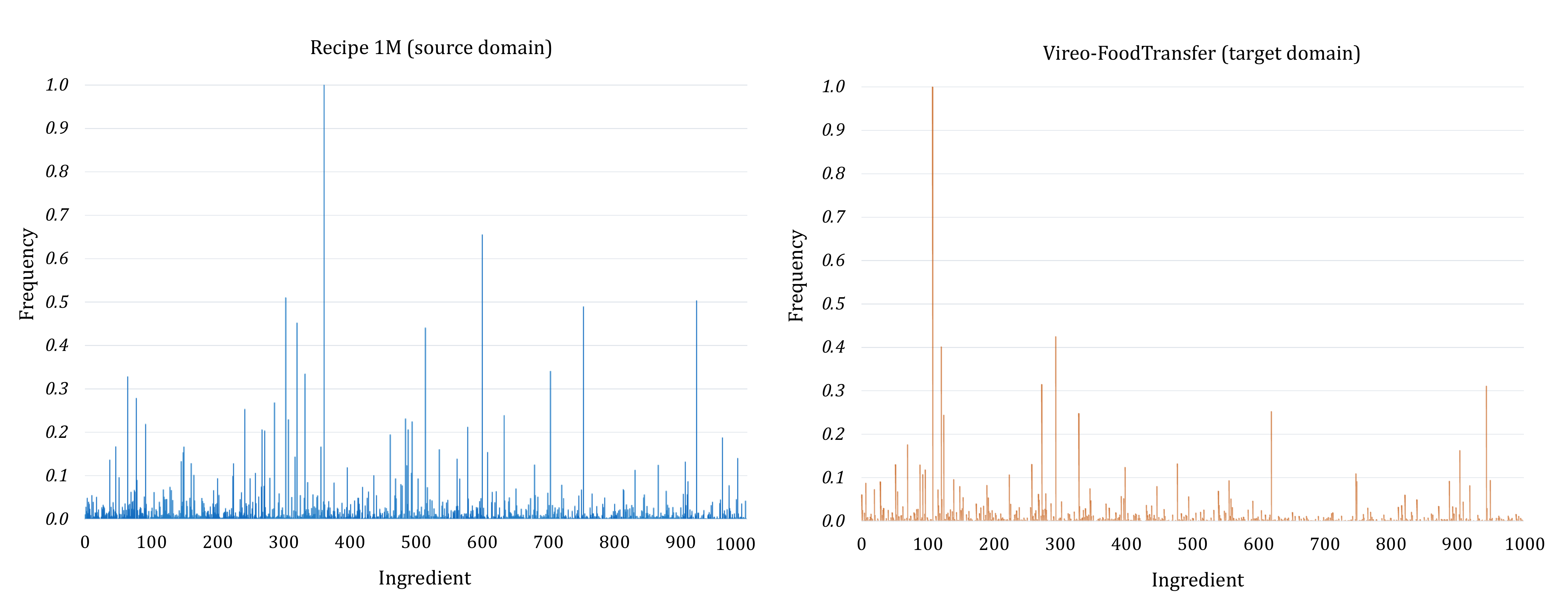}
  \caption{Ingredient usage comparison between Recipe 1M (source domain) and Vireo-FoodTransfer (target domain) datasets. Note that the frequency is normalized within dataset for better comparison.}
  \label{fig:ingredient_usage}
\end{figure*}

Note that the ingredient lists of Vireo-FoodTransfer and Recipe 1M are not fully overlapped. Nevertheless, high-level semantic information (i.e., ingredient labels) is known to be helpful for cross-modal embedding learning as also evidenced in~\cite{JNE, R2GAN, ACME}. To make use of ingredient semantic information and reduce the redundancy of the ingredients, K-Means is employed to obtain a set of 1,000 unified ingredients for the two datasets. Specifically, the Chinese ingredients in Vireo-FoodTransfer are first translated to English. Then clustering is conducted on the Bert ingredient features extracted from the ingredients of Recipe 1M and Vireo-FoodTransfer. For example, ``tomato'', ``tomato puree'' and ``tomato paste'' are clustered into the same class. Among the 1,000 ingredients after clustering, there are 543 common ingredients (e.g., chicken, cucumber and oil) between Recipe 1M and Vireo-FoodTransfer datasets. The numbers of unique ingredients for Recipe 1M (e.g., mayonnaise, philadelphia cheese and nutella) and Vireo-FoodTransfer (e.g., Chinese angelica and lycium barbarum) are 384 and 73 respectively.

Figure~\ref{fig:ingredient_usage} further demonstrates the ingredient usage comparison between Recipe 1M and Vireo-FoodTransfer. Except for the unique ingredients, the frequency of common ingredients in the two datasets vary greatly. For example, the ``soy sauce" is much more heavily used in the Vireo-FoodTransfer while ``toast" is much more common in Recipe 1M. The difference in ingredient usage is one of the main origins of domain gap.

\textbf{Evaluation Metrics.} Similar to~\cite{CCFT}, median rank (MedR) and recall rate at top K (R@K) are adopted as the metrics to evaluate the retrieval adaptation performance. During testing, a subset of
1,000 unseen target images are formed as queries by randomly sampling from the target domain. 
MedR is the median rank of the ground truth recipes for all the queries, and R@K is the percentage value averaged over all the queries at the search depth of K.
The reported MedR and R@K values are the average performance on 10 different sets of target images randomly drawn.

\textbf{Implementation Details.} The backbone of the image encoder is based on ResNet-50~\cite{resnet} pre-trained on ImageNet by replacing the last fully-connected layer with 1024-dimensional output. Similar with~\cite{JNE, CCFT}, the recipe encoder is composed of a Bidirectional LSTM and a hierarchical LSTM for ingredients and instructions encoding respectively. The two features are concatenated with title features and fed into a fully-connected layer to obtain the recipe embedding. The dimension of the Bert features is 768. 
Adam optimizer~\cite{Adam} is adopted for model training with a batch size of 32 in all the experiments. The initial learning rate is set to be 0.0001. The trade-off hyper-parameters $\lambda _1$, $\lambda _2$ and $\lambda _3$ in Equation~\ref{eq:loss} are set to be 0.1, 0.01 and 0.002 respectively. We set the margin $\alpha$=0.3 in Equation~\ref{eq:tri}.

\subsection{Performance Comparison}
The performances of the proposed method against the baseline source-only model, CCFT~\cite{CCFT} and the oracle model are listed in Table~\ref{tab:performance}. Note that the proposed model is essentially built upon CCFT with a recipe mixup block. The source-only and oracle
models are trained by only using the paired data in source and target domains for supervised learning respectively. By directly applying the source model for the target domain with Chinese recipes represented by Bert features, the result is fairly poor, as shown in the first row of Table~\ref{tab:performance}. Instead, we use the language translator~\cite{MT} to translate Chinese recipes to English and then feed into Bert model. The result is boosted sharply as shown in the second row (MedR = 182.0) of the table. In the experiments, we only report results based on the translated recipes. More analysis about the visualization of language gap in recipes is presented in the section~\ref{sec:MT}.

Among the six kinds of RM strategies and two types of RM loss functions, the best performance is achieved by RM$^{s}_4$, which exchanges title and ingredient
sections of source and target recipes (Equation~\ref{eq:rm__source_l2}). Compared with CCFT~\cite{CCFT}, the performance of RM$^{s}_4$
surpasses CCFT with a large margin in terms of MedR and R@K. The MedR is significantly boosted by 13.4 ranks from 128.0 to 114.6, and R@50 is improved by 12.6\% from 30.33 to 34.16.

\begin{table}
  \caption{Image-to-recipe retrieval performance comparison in terms of MedR and R@K. ``E", ``C" and ``C(T)" represents English, Chinese and Chinese-Translated English respectively. RM$^s$ and RM$^{st}$ refer to model training by using recipe mixup losses $\mathcal{L}^s_{rm}$ (in pink) and $\mathcal{L}^{st}_{rm}$ (in yellow) respectively.}
  \label{tab:performance}
  \begin{center}
    \begin{tabular}{l|c|c|c|c}
      \hline
      Method & Domains & MedR & R@10 & R@50\\
      \hline
      \multirow{2}{*}%
      {Source-only} & E$\rightarrow$C & 395.0 & 2.44 & 9.32 \\
      \cline{2-5}
      & \multirow{13}{*}{E$\rightarrow$C(T)} & 182.0 & 10.08 & 26.36 \\
      \cline{1-1} \cline{3-5}
      CCFT~\cite{CCFT} &  & 128.0 & 11.78 & 30.33 \\
      \cline{1-1} \cline{3-5}
      \cellcolor{pink}{RM$^s_1$} & & \cellcolor{pink}{121.2} & \cellcolor{pink}{12.30} & \cellcolor{pink}{32.27} \\
      \cline{1-1} \cline{3-5}
      \cellcolor{pink}{RM$^s_2$} & & \cellcolor{pink}{115.9} & \cellcolor{pink}{14.01} & \cellcolor{pink}{33.61} \\
      \cline{1-1} \cline{3-5}
      \cellcolor{pink}{RM$^s_3$} & & \cellcolor{pink}{125.7} & \cellcolor{pink}{12.08} & \cellcolor{pink}{32.54} \\
      \cline{1-1} \cline{3-5}
      \cellcolor{pink}{\textbf{RM$^s_4$}} & & \cellcolor{pink}{\textbf{114.6}} & \cellcolor{pink}{\textbf{14.42}} & \cellcolor{pink}{\textbf{34.16}} \\
      \cline{1-1} \cline{3-5}
      \cellcolor{pink}{RM$^s_5$} & & \cellcolor{pink}{119.1} & \cellcolor{pink}{14.11} & \cellcolor{pink}{33.13} \\
      \cline{1-1} \cline{3-5}
      \cellcolor{pink}{RM$^s_6$} & & \cellcolor{pink}{120.4} & \cellcolor{pink}{12.53} & \cellcolor{pink}{31.97} \\
      \cline{1-1} \cline{3-5}
      \cellcolor{yellow}{RM$^{st}_1$} & & \cellcolor{yellow}{142.7} & \cellcolor{yellow}{10.32} & \cellcolor{yellow}{28.35} \\
      \cline{1-1} \cline{3-5}
      \cellcolor{yellow}{RM$^{st}_2$} & & \cellcolor{yellow}{141.9} & \cellcolor{yellow}{10.62} & \cellcolor{yellow}{28.49} \\
      \cline{1-1} \cline{3-5}
      \cellcolor{yellow}{RM$^{st}_3$} & & \cellcolor{yellow}{135.3} & \cellcolor{yellow}{11.28} & \cellcolor{yellow}{29.30} \\
      \cline{1-1} \cline{3-5}
      \cellcolor{yellow}{RM$^{st}_4$} & & \cellcolor{yellow}{139.9} & \cellcolor{yellow}{10.79} & \cellcolor{yellow}{29.28} \\
      \cline{1-1} \cline{3-5}
      \cellcolor{yellow}{RM$^{st}_5$} & & \cellcolor{yellow}{142.3} & \cellcolor{yellow}{10.74} & \cellcolor{yellow}{28.24} \\
      \cline{1-1} \cline{3-5}
      \cellcolor{yellow}{RM$^{st}_6$} & & \cellcolor{yellow}{152.6} & \cellcolor{yellow}{9.76} & \cellcolor{yellow}{27.33} \\
      \hline
      Oracle & C(T)$\rightarrow$C(T) & 2.1 & 74.70 & 88.92 \\
      \hline

    \end{tabular}
  \end{center}
\end{table}

\textbf{Why does RM$^s$
work for domain adaptation?} As shown in pink zone of Table~\ref{tab:performance}, all the six RM strategies for recipe mixup loss $\mathcal{L}^s_{rm}$ (Equation~\ref{eq:rm__source_l2}) constantly outperform CCFT in terms of both MedR and R@K. When comparing among single recipe section mixup (i.e., RM$^s_1$, RM$^s_2$ and RM$^s_3$), the ingredient section (i.e., RM$^s_2$) brings the most significant improvement, and title section (i.e., RM$^s_1$) manages to achieve slightly better results than instruction section (i.e., RM$^s_3$). 
When comparing among dual recipe sections mixup (i.e., RM$^s_4$, RM$^s_5$ and RM$^s_6$), 
the performance of RM$^s_4$, which exchanges title and ingredient sections, is superior than RM$^s_5$ and RM$^s_6$. Furthermore, RM$^s_4$ outperforms title-only (RM$^s_1$) and ingredient-only (RM$^s_2$) exchange, which shows that the title and ingredient sections are complementary with each other. Similarly, the combination of title and instructions (RM$^s_5$) exchange also demonstrates superior results compared with RM$^s_1$ and RM$^s_3$. Nevertheless, the property is not applied to the case of ingredient and instruction sections, i.e., RM$^s_6$, which is inferior than RM$^s_2$ but still better than RM$^s_3$. The reasons are two-fold. On the one hand, ingredients demonstrate the composition of a dish which is more informative and discriminative than title and instructions~\cite{salvador2021revamping}. On the other hand, instructions tend to suffer from translation quality of machine translation, due to the naming convention and the free-form writing styles across languages. In contrast, the ingredient section contains the quantity, units and the ingredients, which shows a more universal format across languages, leading to better translation quality.

Compared with CCFT, the RM$^s$ benefits from the constraint imposed by RM loss, which is aware of the shortest geodesic path between source and target domains in constructing the intermediate space. Specifically, the recipe embedding space is regularized to accommodate the mixed recipe features which are artificially generated.
The mixed recipe embeddings are also optimized to maintain a structure obeying the constraint of the shortest geodesic path. In other words, 
during the learning process to construct a desired intermediate domain in the recipe embedding space, the recipe encoder is steered to bridging the domain gap and learn transferable features between source and target domains. This can be validated from the reduced gap between the two domains. Take RM$^s_1$ as an example, compared with CCFT, the distribution distance of source and target recipes (measured by L2 norm with 10K samples) is reduced by 0.19\%. Furthermore, as shown in Table~\ref{tab:performance}, the MedR of RM$^s_1$ is also boosted by 6.8 ranks. This result is also consistent with the distribution distance between target recipe-image pairs (measured by L2 norm with 10K samples), which is reduced by 3.08\%. 

\begin{figure}
  \centering
  \includegraphics[width=0.48\textwidth]{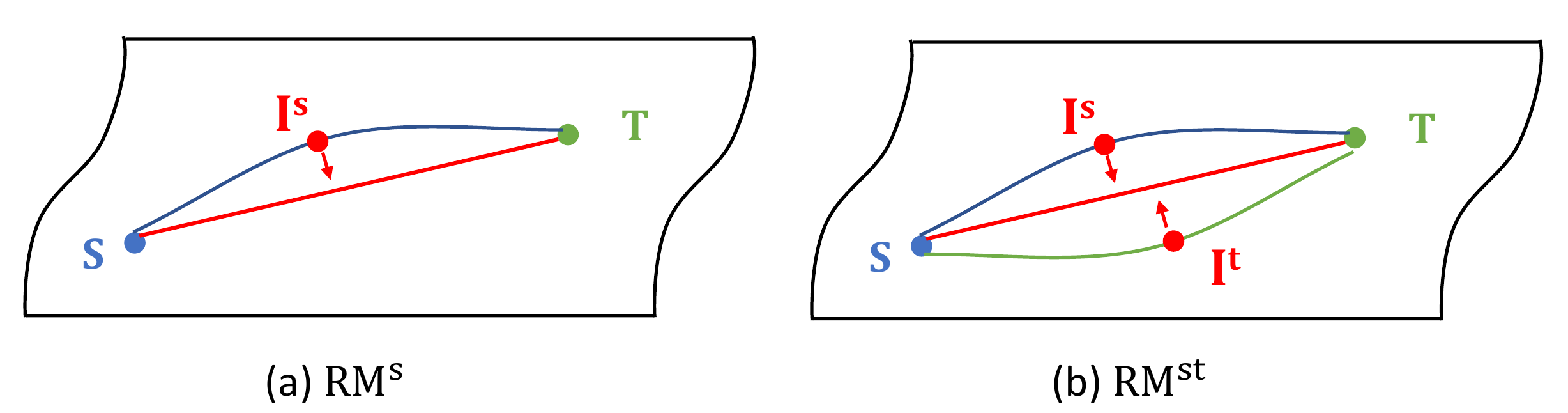}
  \caption{The difference between RM$^s$ and RM$^{st}$. ``S", ``T" and ``I" refer to source, target and intermediate domains respectively. The red line represents the shortest geodesic path in the embedding space. The $I^s$ (in the blue line) and $I^t$ (in the green line) are intermediate domains formed by source and target mixed recipes respectively.}
  \label{fig:intermediandomain}
\end{figure}

\begin{figure*}
  \centering
  \includegraphics[scale=0.272]{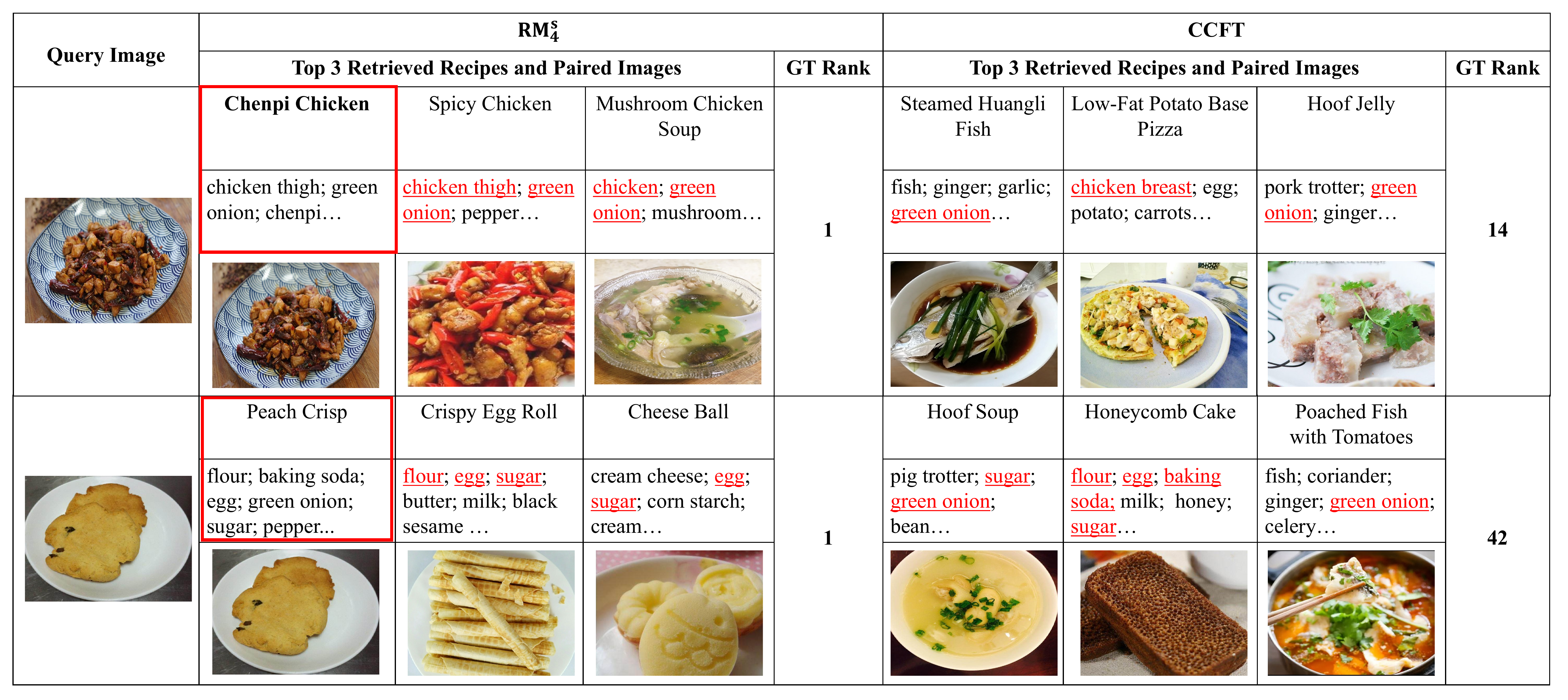}
  \caption{Examples showing the top 3 retrieved results of RM$^s_4$ and CCFT. The ground truth (GT) recipe is highlighted with red bounding box. The common ingredients appeared in the GT recipes are marked in red and underlined.}
  \label{fig:comparison}
\end{figure*}

\begin{figure*}
  \centering
  \includegraphics[scale=0.30]{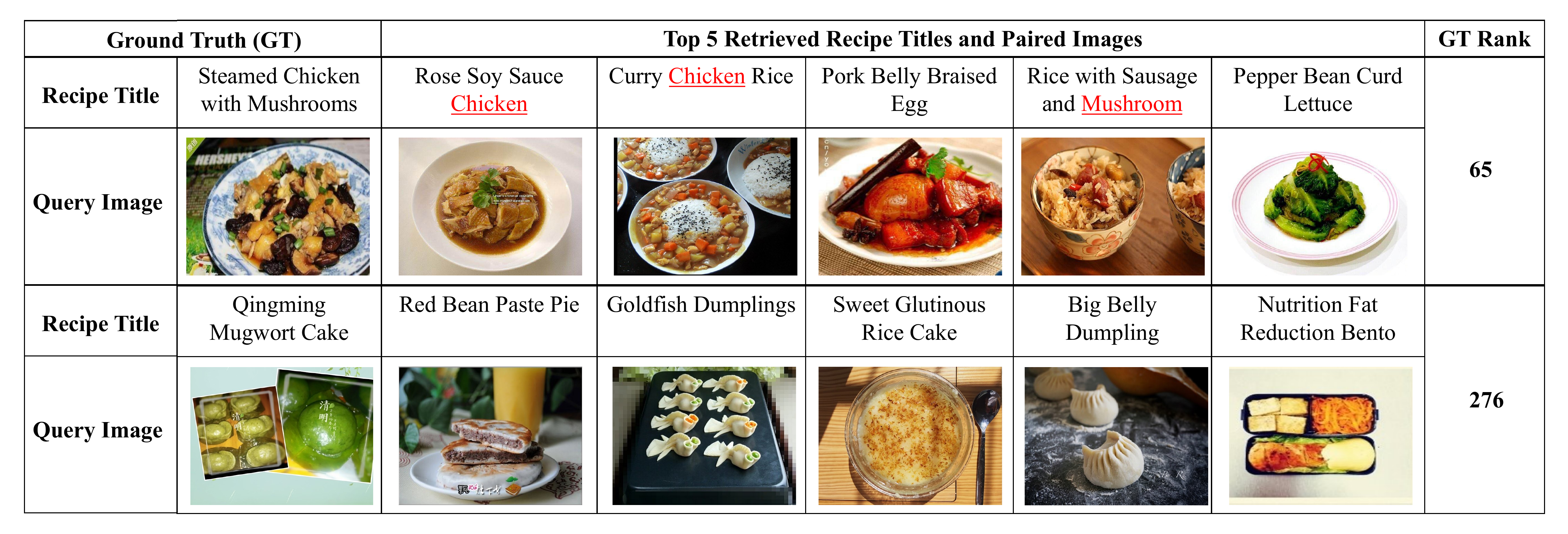}
   \caption{Failure cases showing the top 5 retrieved results of RM$^s_4$.}
  \label{fig:cases}
\end{figure*}

\textbf{Why does RM$^{st}$ does not work?} As RM can produce source and target mixed recipes, RM$^{st}$ investigates the performance by forming the intermediate domain with the two types of mixed recipes, i.e., the model is trained with recipe mixup loss $\mathcal{L}{^{st}_{rm}}$ in the Equation~\ref{eq:rm_l2}. Table~\ref{tab:performance} lists the performance of RM$^{st}$ in yellow zone. Different from RM$^s$, which outperforms CCFT in all the six RM strategies, RM$^{st}$ constantly degrades the performance of CCFT in terms of MedR and R@K. Figure~\ref{fig:intermediandomain} illustrates the difference between RM$^s$ and RM$^{st}$. In fact, the source and mixed recipes intuitively lie in different paths between source and target domains, considering all the sections are opposite with each other. Therefore, extra domain shift is introduced by not only source mixed recipes but also target mixed recipes in RM$^{st}$. The recipe features are obtained by jointly transforming title, ingredients and instructions, while the source and target mixed recipes are symmetric in structure but distinctive in contents. As a consequence, the objectives of RM$^{st}$ forcing source and target mixed recipes to stay on the shortest geodesic path in the embedding space simultaneously are not necessarily aligned.
In other words, with two different objectives to fulfill the shortest geodesic path constraint, RM$^{st}$ results in negative transfer.

\textbf{Qualitative results.} As shown in Figure~\ref{fig:comparison}, two typical examples are presented showing the top 3 retrieved recipes of RM$^s_4$ and CCFT~\cite{CCFT}. Note that the recipes shown in the figure is originally Chinese and translated into English for presentation purpose. Only title and major ingredients are presented for a recipe to save space. In both examples, RM$^s_4$ manages to rank the ground truth (GT) recipes in the first place while the ranks of CCFT are worse at the positions of 14 and 8 respectively. Furthermore, the recipes in the top 3 of RM$^s_4$ also demonstrate more overlapped ingredients with the GT recipes. For instance, in the first example (rows 3-5), the second (``Spicy Chicken") and third (``Mushroom Chicken Soup") retrieved recipes of RM$^s_4$ contain the major ingredients ``chicken thigh" or ``chicken" and ``green onion". In contrast, CCFT only contains one of ``green onion" or ``chicken breast". The result shows that RM$^s_4$ manages to capture more discriminate features and thus achieves better retrieval performance. 

Nevertheless, the overall retrieval performance of RM$^s_4$ is much worse than the traditional retrieval models~\cite{ACME, salvador2021revamping} and domain adaptation models~\cite{CCFT} with mono-lingual recipes. Indeed, cross-lingual adaptation for recipe retrieval is an extremely challenging problem. Figure~\ref{fig:cases} shows two standard failure cases of the top 5 retrieved recipes by RM$^s_4$, where the ranks of GT recipes are in relatively large positions. In the first example (rows 2-3), although the GT recipe is ranked in 65, 3 out of the top 5 recipes contain common main ingredients with the GT recipe, for example, ``chicken" and ``mushroom". In the second example (rows 4-5), the query image (``Qingming Mugwort Cake") is one kind of special food in China which is particularly made in a traditional Chinese Qingming festival, as a result, it is not surprising that the rank for the GT recipe is as bad as 276. The reason is that little common knowledge can be transferred from Recipe 1M where majority of the recipes are quite different western food. Interestingly, although the visual appearance of the top 5 images and the query image is quite different, four of the recipes are made from ``flour", which is the same with the GT recipe.

\begin{figure*}
  \centering
  \includegraphics[scale=0.38]{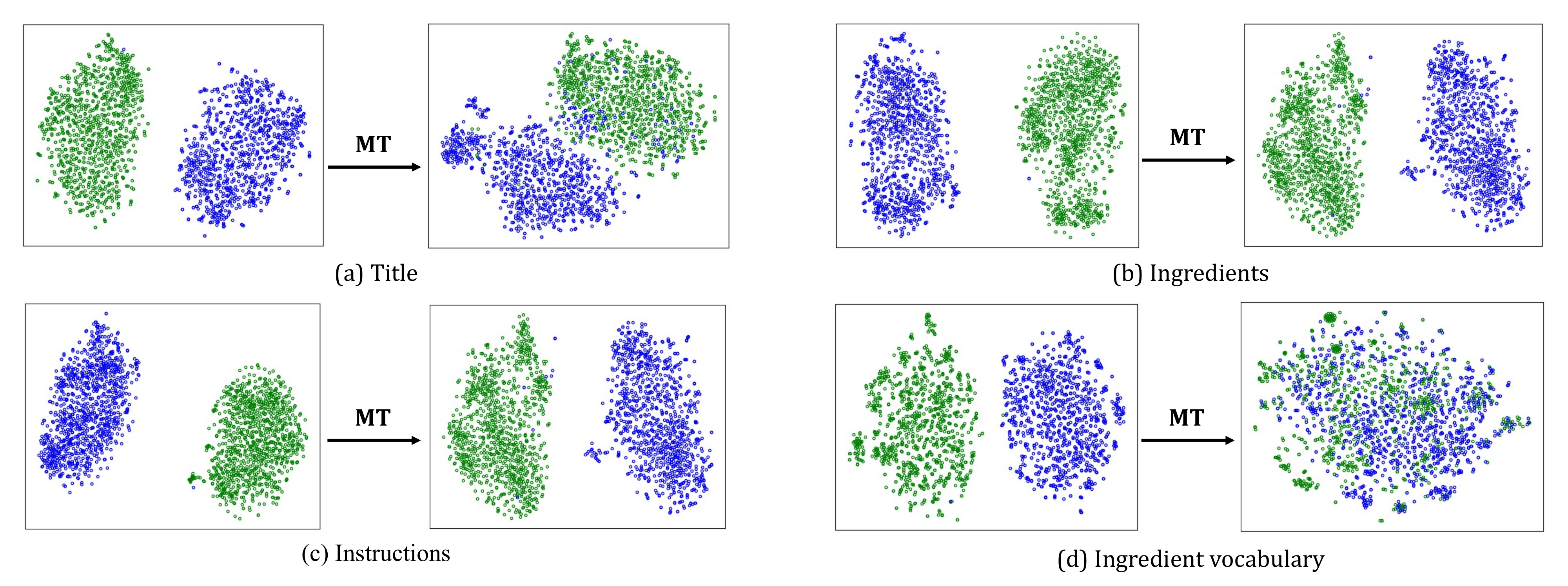}
  \caption{Visualization of recipe Bert features for title, ingredients and instructions and ingredient vocabulary by t-SNE  on source (Green dots) and target (Blue dots) domains before and after machine translation (MT).}
  \label{fig:visualization}
\end{figure*}

\subsection{Impact of Machine Translation}
\label{sec:MT}
Figure~\ref{fig:visualization} visualizes the Bert features of three recipe sections and ingredient vocabulary in the source (Recipe 1M) and target (Vireo-FoodTransfer) by t-SNE~\cite{t-SNE} before and after machine translation (MT). As ingredient and instruction sections contains multiple sentences, the visualization results are based on the average of all the Bert features for each section. Different from the recipe sections, ingredient vocabulary refers to the post-processed ingredient labels (each ingredient label includes a few words, e.g, olive oil, white sugar and mushroom) in each dataset, specifically, the 4,102 ingredients in Recipe 1M and 1,635 ingredients in Vireo-FoodTransfer.
 Observed from the results before machine translation (left side of each sub-figure in Figure~\ref{fig:visualization}), all the multilingual Bert features from source and target domains are clearly separated into two distinctive sets, including the ingredient vocabulary. It shows that the domain gap between source and target domains is quite evident. Furthermore, the ``inherent" translation by multi-lingual Bert model fails to capture the same semantics across different languages. However, after explicitly translating the Chinese recipes to English (right side of each sub-figure in Figure~\ref{fig:visualization}), it can be observed that the distance between source and target domains is shorten, i.e, the domain gap is reduced. In particular, the reduced gap of ingredient vocabulary are much more evident than all the three recipe sections. The result is not surprising because the ingredient vocabulary is pre-processed into ingredient labels beforehand, therefore, some of the common ingredients of the Chinese-translated English in Vireo-FoodTransfer can be exactly the same with the original English ingredients in Recipe1M. 

\subsection{Ablation Study}

\begin{table}
  \caption{Ablation study based on RM$^s_4$.}
  \label{tab:ablation}
  \begin{center}
    \begin{tabular}{l|c|c|c}
      \hline
      Methods & MedR & R@10 & R@50 \\
      \hline
      RM$^s_4$ w/o adv & 124.7 & 12.18 & 31.70\\
      \hline
      RM$^s_4$ w/o sem & 119.5 & 13.78 & 32.65\\
      \hline
      RM$^s_4$ w/o gen & 117.8 & 13.95 & 32.92\\
      \hline
      RM$^s_4$ w/o rm (CCFT~\cite{CCFT}) & 128.0 & 11.78 & 30.33\\
      \hline
      RM$^s_4$ (full) & 114.6 & 14.42 & 34.16\\

      \hline

    \end{tabular}
  \end{center}
\end{table}

The significance of each RM$^s_4$ component is assessed. Table~\ref{tab:ablation} lists the performances of the ablation models with one of the loss functions being excluded from training in turn.
Adversarial learning (adv) shows a high impact on the performance, where MedR and R@K drop dramatically without employing domain discriminator for feature alignment. In contrast, the semantic regularization (sem) for ingredient recognition 
and image generation (gen) from recipe embedding have a lower impact on the performance. Without sem, both MedR and R@K degrade a bit. Similarly, the performance
also drops slightly without recipe-to-image generation. The result is aligned with CCFT~\cite{CCFT}, where ingredient recognition from image embedding is more important than image generation for the retrieval performance. 
Finally and the worst, when RM is taken away, the largest margin of drop is noticed. Since RM$^s_4$ is built upon CCFT with recipe mixup, the performance without RM is the same with CCFT.

\section{Discussion}
We recapitulate and discuss the limitations of this paper in three-fold.
First, there is a performance gap between the proposed model and the oracle model which is trained with paired data in the target domain. The performance gap is also larger than the result reported in CCFT~\cite{CCFT} on the three Asian cuisines without language gap. Second, the proposed recipe mixup has significantly boosted the performance of traditional model (without domain adaptation) and the recent CCFT (with domain adaptation for the recipes in the same language). Despite these achievements, the physical meaning of using recipe mixup for domain augmentation remains not being fully understood. Similar studies also point out the issue and some progress has been made to explain mixup~\cite{mixup18, mixup21}. Third, we notice that there is an apparent performance gap between using multi-lingual Bert for ``inherent" translation and using off-the-shelf language translator for explicit translation. Exploring internal properties of languages (e.g., lexical similarity or structural similarities)~\cite{karthikeyan2019cross} and fine-grained alignment (e.g., subword)~\cite{bojanowski2017enriching} between different languages could be one possible solution to ease the dependency of machine translation.

\section{Conclusion}
We have presented a novel recipe mixup method for cross-lingual adaptation in the context of image-to-recipe retrieval, which outperforms the baseline models with a large margin. Through the empirical experiments, we have shown that all six kinds of recipe mixup strategies with source mixed recipes constantly achieve better performance than the baseline models. Exchanging both title and ingredient sections between source and target recipes attains the best performance. By using both the source and target mixed recipes as intermediate domain, negative transfer appears. The future work includes exploration of recipe modeling using transformers and machine translation free cross-lingual adaptation.

\section*{Acknowledgments}
The work described in this paper was partially supported by a grant from the Research Grants Council of the Hong Kong Special Administrative Region, China (CityU 11203517), a CityU MF\_EXT grant (project no. 9678180), and a Shanghai Pujiang Program (20PJ1401900).

\bibliographystyle{ACM-Reference-Format}
\bibliography{RecipeMixup}

\end{document}